\documentclass{article}

\usepackage{cite}
\usepackage{amsmath,amssymb,amsfonts}
\usepackage{algorithmic}
\usepackage{graphicx}
\usepackage{textcomp}
\usepackage{xcolor}
\def\BibTeX{{\rm B\kern-.05em{\sc i\kern-.025em b}\kern-.08em
    T\kern-.1667em\lower.7ex\hbox{E}\kern-.125emX}}
\usepackage[hidelinks]{hyperref}       % hyperlinks
\usepackage{listings}
\usepackage{booktabs}       % professional-quality tables
\usepackage{multirow}
\usepackage{bm}

\usepackage{arxiv}
\usepackage[utf8]{inputenc} % allow utf-8 input
\usepackage[T1]{fontenc}    % use 8-bit T1 fonts
\usepackage{url}            % simple URL typesetting
\usepackage{nicefrac}       % compact symbols for 1/2, etc.
\usepackage{microtype}      % microtypography
\usepackage[square,numbers]{natbib}
\usepackage{doi}
\usepackage{dblfloatfix}
\usepackage{caption}

\definecolor{codegreen}{rgb}{0,0.6,0}
\definecolor{codegray}{rgb}{0.5,0.5,0.5}
\definecolor{codepurple}{rgb}{0.58,0,0.82}
\definecolor{backcolour}{rgb}{255,255,255}

\lstdefinestyle{mystyle}{
    backgroundcolor=\color{backcolour},   
    commentstyle=\color{codegreen},
    keywordstyle=\color{blue},
    numberstyle=\tiny\color{codegray},
    stringstyle=\color{codegreen},
    basicstyle=\ttfamily\scriptsize,
    breakatwhitespace=false,         
    breaklines=true,                 
    captionpos=b,                    
    keepspaces=true,                 
    numbers=none,                    
    numbersep=5pt,                  
    showspaces=false,                
    showstringspaces=false,
    showtabs=false,                  
    tabsize=2,
    mathescape=true
}

\lstdefinelanguage{Modelica}{%
  language     = C++,
  alsoletter={.},
  morekeywords = {equation,model,parameter,constant,
  	package,end,annotation, partial, 
  	replaceable, then, redeclare, final,
  	extends},
  	morecomment=[n][\color{red}]{Modelica.}{ },
  emph={assert,abs,Real,
    Modelica.Media.Interfaces.PartialMedium,
    Modelica.Fluid.Interfaces.FluidPort_a,
    Modelica.Fluid.Interfaces.FluidPort_b,
    Medium_a,
    Medium_b,
    PartialSaturatedControlVolume
  	},
  	emphstyle=\color{red}	
}

\renewcommand{\headeright}{ES2026 (Preprint)}
\renewcommand{\undertitle}{\textbf{Preprint} for the ASME 2026 20th International Conference on Energy Sustainability \\
July 26--29, 2026, Bellevue, WA}
\renewcommand{\shorttitle}{\emph{A. Mensikova et al.: Mapping AI in LCA Using LLMs}}
\date{February 25, 2026}

\hypersetup{%
  pdftitle  = {Mapping the Landscape of Artificial Intelligence in Life Cycle Assessment Using Large Language Models},
  pdfauthor = {Anastasija Mensikova, Donna M. Rizzo, Kathryn Hinkelman},
  pdfsubject = {The ASME 2026 20th International Conference on Energy Sustainability (ES2026)},
  colorlinks,
  linkcolor=black,
  urlcolor=black,
  citecolor=black,
  pdfpagelayout = SinglePage,
  pdfcreator = pdflatex,
  pdfproducer = pdflatex}

\begin{document}

\title{Mapping the Landscape of Artificial Intelligence in Life Cycle Assessment Using Large Language Models}

\author{ 
\href{https://orcid.org/0000-0001-7448-283X}{\includegraphics[scale=0.06]{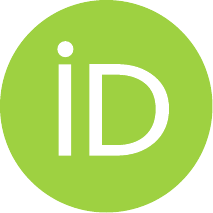}\hspace{1mm}
    Anastasija~Mensikova} \\
	University of Vermont\\
	Burlington, VT, USA \\
    \texttt{anastasija.mensikova@uvm.edu} \\
\And
    \href{https://orcid.org/0000-0003-4123-5028}{\includegraphics[scale=0.06]{orcid.pdf}\hspace{1mm}Donna~M.~Rizzo} \\
	University of Vermont\\
	Burlington, VT, USA \\
	\texttt{donna.rizzo@uvm.edu} \\
\And
    \href{https://orcid.org/0000-0002-8297-6036}{\includegraphics[scale=0.06]{orcid.pdf}\hspace{1mm}Kathryn~Hinkelman} \\
	University of Vermont\\
	Burlington, VT, USA \\
	\texttt{kathryn.hinkelaman@uvm.edu} \\
}

\maketitle

\begin{abstract}
Integration of artificial intelligence (AI) into life cycle assessment (LCA) has accelerated in recent years, with numerous studies successfully adapting machine learning algorithms to support various stages of LCA. 
Despite this rapid development, comprehensive and broad synthesis of AI-LCA research remains limited. 
To address this gap, this study presents a detailed review of published work at the intersection of AI and LCA, leveraging large language models (LLMs) to identify current trends, emerging themes, and future directions. 
Our analyses reveal that as LCA research continues to expand, the adoption of AI technologies has grown dramatically, with a noticeable shift toward LLM-driven approaches, continued increases in ML applications, and statistically significant correlations between AI approaches and corresponding LCA stages. 
% \textcolor{red}{[one sentence on how effective LLM-based methods were for mapping AI and LCA (lit review)]}
By integrating LLM-based text-mining methods with traditional literature review techniques, this study introduces a dynamic and effective framework capable of capturing both high‑level research trends and nuanced conceptual patterns (themes) across the field.
% From improving data collection practices to aiding the understanding of the resulting report, AI and ML technologies are successfully improving lives of LCA practitioners.
% This analysis further emphasizes that the environmental burdens of AI, especially LLMs, must be explicitly considered when integrating these tools into LCA practice. 
Collectively, these findings demonstrate the potential of LLM-assisted methodologies to support large-scale, reproducible reviews across broad research domains, while also evaluating pathways for computationally-efficient LCA in the context of rapidly developing AI technologies. 
In doing so, this work helps LCA practitioners incorporate state-of-the-art tools and timely insights into environmental assessments that can enhance the rigor and quality of sustainability-driven decisions and decision-making processes.

\end{abstract}


%%%%%%%%%  BODY OF PAPER %%%%%%%%%%%%%%%%%%%%%%%%%%%%%%%%%%%%%%

\section{Introduction} 

% In the age of Artificial Intelligence (AI) and sustainable energy transitions, Life Cycle Assessment (LCA) provides critical knowledge for environmentally-informed design and operational decisions. 
As Artificial Intelligence (AI) becomes increasingly integrated into modern technological and industrial systems, the need for environmentally responsible innovation is more important than ever. 
Life Cycle Assessment (LCA), governed by ISO‑14040~\cite{iso14040}, remains the leading framework for quantifying comprehensive environmental impacts across the entire life cycle of products, processes, and systems, from resource extraction to end-of-life.
% Environmental impact categorizes span human and ecological health, such as carcinogenic effects, global warming, eutrophication, and resource depletion.  
% There are four main stages: goal and scope definition, life cycle inventory (LCI), life cycle impact assessment (LCIA), and interpretation.
% This provides an invaluable framework for making design and operational decisions on an environmental basis
Meanwhile, AI has emerged as a major change maker across numerous fields and industries, introducing not only new environmental and energy challenges~\cite{Strubell2019}, but also opportunities for systematic efficiency gains and transformational innovations~\cite{Rolnick2019}. 
AI systems themselves pose environmental burdens, and LCA must evolve to assess them. 
Simultaneously, AI offers tools that can accelerate or reshape LCA itself.
Despite growing attention to the intersection of AI and LCA, questions remain on how LCA as a practice will be improved in the era of generative AI.
%~\cite{Nwagwu2025}. 

% Despite growing attention to the intersection of AI and LCA, it is unclear how AI can best support the LCA process.
% and how LCA can best allocate AI effects in case studies.
Existing studies highlight the integration of AI within LCA across many different applications and methodological stages. 
Several use machine learning (ML) and large language models (LLMs) to support specific LCA tasks~\cite{Balaji2025}, in particular data completion within life cycle inventory (LCI) and report interpretation~\cite{Comago2023}. 
Other works focus on domain-specific applications, such as wastewater treatment systems~\cite{IbnMohammed2023} and sustainable manufacturing~\cite{Preuss2024},
showcasing the potential of AI-enabled methods to improve predictions and inform decisions in specific applications. 
%, construction materials~\cite{Koyamparambath2022}
More recent contributions have also examined the broader implications of integrating LLMs into LCA workflows, including challenges related to data consistency and computational cost. 
For example, Jiang et al.~\cite{Jiang2024} analyze the life-cycle energy and carbon impacts of LLM-powered systems. 
Although these challenges are crucial to address when utilizing LLMs, their ability to process unstructured literature at scale and automate interpretation tasks is what makes them great at assisting LCA tasks.
The duality of AI supporting LCA, and LCA evaluating AI’s environmental impacts is therefore especially important to highlight.

Although these contributions illustrate the growing role of AI in LCA research, most centered on individual case studies, specific sectors, or narrowly defined methodological tasks. 
As a result, two critical research gaps remain. 
First, there remains a need for comprehensive, data-driven understanding of how AI has been used in LCA research and how these applications have evolved over time. 
Second, LLM-assisted literature reviews have demonstrated success in quickly producing deep and accurate insights~\cite{Scherbakov2025}, but these methods have not yet been implemented in the LCA domain, to our knowledge.
Instead, previous reviews have been conducted through systematic manual methods~\cite{Preuss2024,IbnMohammed2023}.
Thus, questions remain about which LLMs are most effective (in terms of both accuracy and computational costs) and how prompts should be structured in LCA applications. 
Addressing these gaps, this work combines embedding-based clustering with targeted LLM–assisted interpretation to systematically map the landscape of AI applications in LCA, while simultaneously demonstrating a scalable methodology that supports human-directed literature reviews.
Specifically, we aim to understand how using lightweight open-source language models can provide an effective, yet energy-conscious, way to explore trends in technological advancements of LCA methodologies and therefore provide a comprehensive overview of the trajectory of the field.
% Specifically, we aim to understand... \textcolor{blue}{[Anastasija -- can you finish this sentence? (assuming you can think of a goal that doesn't repeat what we already said).]}

% The intersection of LCA and AI is growing rapidly, which has been summarized in previous reviews.
% These reviews primarily examine the use of ML and AI to support specific LCA tasks, discuss opportunities and limitations of AI-enabled LCA workflows, or focus on challenges associated with data availability and model interpretability \cite{Preuss2024,IbnMohammed2023}. % is this okay? - Good sentence, but I'm not sure how to integrate it. I took the citation above for now, and we can expand the background lit review more later if requested.
% While previous literature reviews in LCA have typically been conducted through systematic manual methods or using simplified automation tools, this work provides not only a comprehensive overview of AI applications within LCA, but also a novel methodology for streamlining human-directed literature reviews with transformer-based architectures. 

The rest of this paper is organized as follows. 
Integrating LLM-based approaches into a systematic literature review structure, Section~\ref{sec:methods} presents the methods, including both wide-reaching metadata analysis and deep contextual analysis. 
Section~\ref{sec:results} presents the results, providing a comprehensive overview of the findings via model outputs and visualizations.
After conclusions in Section~\ref{sec:conclusion}, a link to the metadata repository for this study is provided.

%%%% Methods %%%%%%%%%%%%%%%%%%%%%%%%%%%%%%%%%%%
\section{Methods} \label{sec:methods}

The workflow consists of four main stages: (1) data collection, (2) data preparation, (3) metadata analysis, and (4) content synthesis. 
Through this process, the goal is to map the AI landscape of original LCA research while evaluating LLM-based approaches for systematic reviews.
As such, several steps contain both LLM-based and non-LLM procedures to validate and compare processes and outputs.
The specific methodological details are as follows.

\subsection{Data Collection}

\begin{figure}[htbp]
    \centering
    \includegraphics[width=0.5\linewidth]{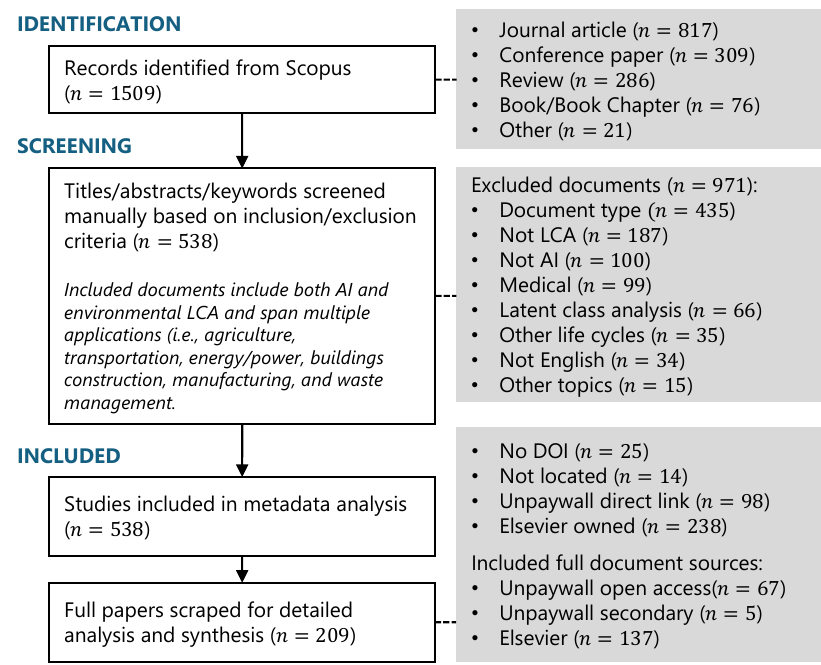}
    \caption{Review protocol to identify, screen, and collect documents based on PRISMA 2020~\cite{PRISMA}. }
    \label{fig:PRISMA}
\end{figure}

This systematic data collection and screening process follows the PRISMA Standard~\cite{PRISMA}, as summarized in Fig.~\ref{fig:PRISMA}. 
First, candidate papers were identified following a refined Scopus search across title, abstract, and keyword field. 
Several search term combinations were evaluated to capture the variety of LCA/AI terminologies with minimal pre-filtering.
Ensuring accuracy and relevancy to convergent original research on AI and LCA, the final search criteria was (\texttt{life cycle assess*, life cycle analy*, life cycle inventory, life cycle impact assess*, LCA, LCI, OR LCIA) AND (machine learning, artificial intelligence, large language model*, ML, AI, OR LLM}). This search resulted in 1509 documents.

Second, each document's metadata underwent a screening process, which involved a manual read-through of the abstracts and titles.
Exclusion criteria included document type, language, or major topical domains.   
Records excluded based on document type were not original research, not individual papers (i.e., conference proceedings list), or missing abstracts. 
Several papers used LCA or AI as acronyms but referred to other topics, such as \emph{latent class analysis}, \emph{aggregation index}, or numerous medical abbreviations.
This screening process resulted in 538 papers deemed relevant to the topic of AI within LCA, all of which were included in some of the meta-data analysis.

Next, because full texts are critical for leveraging LLMs in exploratory analysis, we queried research databases to automatically extract as many full text as possible. 
First, only 513 papers included a DOI, a unique identifier critical for successful data mining and further analysis of the papers. 
\textit{Unpaywall}\cite{Piwowar2018}, a database containing open-access papers, was utilized as the first step of the data mining process. 
Queries were made via its Python library, \textit{unpywall} (v0.2.3), which acts as a conduit to \textit{Unpaywall}'s REST API. Due to the restrictive nature of published research, only \textit{72} papers were identified as open-access via \textit{Unpaywall}, and their complete PDF files were successfully scraped. 

To continue building the full-text corpus of this analysis, out of the remaining \textit{441} papers, \textit{238} belonged to \textit{Elsevier} journals by selecting DOIs starting with \textit{10.1016}, a common \textit{Elsevier} publication identifier. 
Calls were made to \textit{Elsevier API}, querying for full text, when available. 
This procedure yielded \textit{137} xml documents, containing metadata and full texts of the papers. 
However, it is important to note that the structure of the xml documents was not uniform, which meant that additional cleaning and advanced filtering was required to extract complete full text.
Other APIs, such as \textit{Springer Nature}, were also queried, but they proved to be unsuccessful in retrieving full text due to subscription limitations. 
All together, \textit{209} successfully retrieved papers were kept as the final and complete dataset for the full-text content synthesis.

\subsection{Data Preparation}

To ensure accuracy of results, the next step of the process involved cleaning and preparing retrieved text for analysis. 
Both the originally downloaded Scopus abstracts and the newly scraped full texts were used in the study and therefore underwent cleaning, traditionally applied to text prior to using natural language processing tools. 
We performed metadata analysis on both the 538 and 209 sets of studies to ensure the smaller sampling population reflects the larger trends.
Due to the nature of the algorithms used throughout this research, the cleaning and preparation performed were rather simple. 
Simple removal of special characters, line breaks and extra spaces was sufficient. 
For algorithms used in this analysis, additional pre-processing steps, such as stemming and lemmatization, were not necessary.
Instead, preserving the original structure and therefore semantic meaning of sentences was critical for the purpose of training embeddings.

A similar procedure was followed when preparing full text for subsequent LLM analysis. 
While text from PDF documents scraped via Unpaywall were simply extracted with PDFReader, xml documents from Elsevier required a separate extraction process. 
The lack of uniformity among xml documents presented a challenge, which was addressed with adaptable tag pattern recognition throughout the document extraction process. 
Overall, the process concluded successfully, but the potential for messy data was taken into consideration when parsing and analyzing the data. 
The full text of all extracted papers was subsequently cleaned by removing special characters and academic identifiers (URLs, emails, DOIs), reconstructing broken words, and normalizing the overall structure. 
This process ensured that the resulting text was clean and clear, providing a strong platform for the LLM analysis.

\subsection{Metadata Analysis}

The first step in exploring general research trends within LCA and AI research was clustering papers ($n=209$). 
% Clustering is an unsupervised machine learning method used to partition unlabeled data points into groups, and in the case of this analysis there are 209 papers representing the data points. 
% Paper DOIs are used to identify each paper throughout the analysis. 
Because clustering using full text would introduce too much noise and potentially lead to overfitting, only abstracts were used at this stage. 
However before building the clusters, papers abstracts were converted into numerical representations, or embeddings. 
A pre-trained Sentence-BERT transformer model, \textit{all-MiniLM-L6-v2}, was used to generate these dense numerical vectors because it is designed for semantic similarity, is lightweight and stable, works well in scientific contexts, and captures meaning rather than simple term overlap. 
The result is 209 384-dimensional vectors, each representing one abstract.

Although it is possible to perform clustering on raw embeddings, 384-dimensional vectors can be quite noisy and fall victim to the curse of dimensionality. 
It is therefore imperative to use dimensionality reduction prior to clustering to maximize the accuracy and overall performance of the algorithm. 
For this, Uniform Manifold Approximation and Projection (UMAP) was chosen as the dimensionality reduction method, which projects a high-dimensional embedding space into a lower-dimensional representation while preserving local semantic structure. 
Unlike other methods of dimensionality reduction, UMAP preserves local structure, is more adaptable to nonlinear relationships, and is commonly recommended for optimized performance in unsupervised clustering \cite{mcinnes2018umap}. 
UMAP configuration was specifically selected for this task in order to preserve local semantic relationships among abstracts, which is essential for distinguishing fine‑grained topical clusters.
For this analysis, UMAP configured with 10 neighbors, 10 components, a cosine distance, and a minimum distance of zero to preserve fine‑grained semantic structure while allowing similar papers to form tightly clustered neighborhoods. 
% to emphasize local structure in the embedding space, which is critical for identifying fine-grained thematic groupings within a heterogeneous literature corpus. 
% The dimensionality of the reduced space was set to 10 components in order to balance noise reduction while preserving semantic structure. 
% Cosine distance was chosen as the similarity metric to remain consistent with the geometry of the sentence embeddings. 
% A minimum distance of zero was chosen to allow similar papers to form tightly packed neighborhoods and therefore improve the performance of subsequent clustering. 
Lastly, a fixed random state was used to ensure reproducibility.

Once embeddings were trained and their dimensionality was reduced, they were passed to HDBSCAN 
% Although there exists a large variety of clustering algorithms, due to the nature of the problem at hand, HDBSCAN 
(Hierarchical Density-Based Spatial Clustering of Applications with Noise)~\cite{hdbscan} for clustering.
% was selected as the most fitting way to group papers into topics. 
This clustering algorithm perfectly suited the exploratory data analysis of interest -- it does not require a specified number of clusters; it is able to naturally handle uneven cluster density and explicitly specify outliers; and it works well with embedding spaces. 
These features are particularly important for AI–LCA literature, which is heterogeneous, interdisciplinary, and contains a substantial number of methodological outliers. 
In HDBSCAN, the minimum cluster size was set to 15 to avoid forming clusters around very small groups of papers and to focus the analysis on stable and well-represented themes. 
The parameter \texttt{min\_samples} was set to one to allow cluster formation around dense local neighborhoods while still permitting the identification of noise points. 
Since UMAP transformed the data into a Euclidean geometric space, Euclidean distance was used as the clustering metric, and the \texttt{leaf} cluster selection method was chosen to capture finer-grained substructure in the literature. 

Upon creating the clusters, it was crucial to expand and analyze them to both verify the accuracy of the clustering algorithm and get an understanding of the themes available in the dataset. 
While a quick manual skim through the clusters and their abstracts gave a confirmation that the algorithm did indeed separate the dataset into clear themes, term extraction was the second natural step to getting a better overview of each cluster. 
% It is important to note that papers identified as outliers (\texttt{cluster = -1}) were not considered as part of further cluster analysis to avoid introducing noise to the themes being explored. 
Excluding outliers, the first method used for extracting cluster keywords was Term Frequency-Inverse Document Frequency (TF-IDF), which is a traditional method for measuring a word's importance by combining how often a term appears with how uncommon it is across documents. 
To do this, all abstracts within each cluster were collected; TF-IDF values were computed and averaged across papers; and the top terms were returned. 
To refine the results further, noun phrases were extracted with \texttt{spaCy}, filtered to removed noisy and domain-generic terms, and then re-ranked with TF-IDF so that  
% additional steps were added to the keyword extraction process. 
% First, upon organizing abstracts into clusters, \texttt{spaCy} was used to extract noun phrases. 
% To avoid unnecessary noise, generic academic terms, publisher information, and domain-generic terms were all removed from those phrases. 
% The remaining phrases were then re-ranked using the same TF-IDF method as before, and 
the top multi-word phrases per cluster provided a clearer overview of each group.

To provide an alternative and more in-depth look at each cluster, LLM-assisted cluster interpretations were subsequently developed. 
Using HDBSCAN's probability rankings for each paper, the top 15 abstracts were selected from each cluster to reduce noise and minimize the risk of LLM hallucinations. 
A lightweight open-source instruction-following large language model (\texttt{LLaMA-3 8B}) was chosen for cluster interpretation due to its strong performance in constrained summarization tasks and its ability to be run locally with reproducible behavior. 
The model was used at low temperature (\texttt{0.1}) to generate concise and consistent cluster labels without influencing the clustering process itself. 
The most crucial and challenging step in ensuring a successful LLM output was prompt engineering, which involved going through multiple iterations to create a prompt specific enough to gather required information but simple enough to avoid confusion and hallucinations. 
The resulting prompt requested the model to read through all abstracts of the cluster and return exactly three lines: one with a title to the cluster, one with a brief description of the cluster, and one giving a brief explanation of how AI is implemented in papers of this cluster. 
It was critical to explicitly request the model to be as specific as possible and avoid obvious, generic terminology when generating titles such as \texttt{Life Cycle Assessment}.

With the selected methodology, it is worth noting that the use of LLMs raises important questions regarding computational cost and energy consumption. 
Recent studies highlight how prompt design and inference strategies specifically can influence energy consumption~\cite{Rubei_2025}. 
In this study, LLM usage was intentionally constrained through concise, structured prompts and lightweight models, reducing inference overhead while still enabling effective analysis of the LCA-AI research landscape.

Finally, to compare AI-based methods with standard statistical approaches, we also analyzed metadata using the CorTexT Manager~\cite{CorTexT}, a free online platform for analyzing textual datasets.
After importing metadata into CorTexT ($n=538$), term extraction algorithms identified the top 500 terms across titles, abstracts, and keywords based on their specificity using $\chi^2$ and occurrence frequency, long-established standards~\cite{kageura1996}.
Terms from two to four words long were considered.
Then, terms were manually categorized and grouped based on (1) LCA terminology, (2) AI terminology, and (3) energy/power terminology. 
Several visualization options were explored to understand the evolution of various topics over time and the statistical correlations between them.
This includes contingency matrix visualizations, which were used to map the joint distribution across AI and LCA topics. 
Deviation measures based on $\chi^2$ can indicate positive correlations (e.g., a score of 3 indicates a 300\% higher than expected correlation between terms), or negative correlations (e.g., -3 indicates the number of expected co-occurrences is 300\% higher than the observed)~\cite{CorTexT}. 
Statistical significance was set at $p<0.05$.

\subsection{Content Synthesis}

Looking at abstracts and their distribution across clusters was imperative in aiding the analysis and understanding of the role AI plays in LCA research. 
However, in order to understand what specific AI/ML technologies are the most prominent in LCA or, for example, which LCA stages are receiving the most attention in research, it was imperative to explore full text of papers. 
Full documents were therefore analyzed for trends in technological advancements within LCA-AI research to identify specific algorithmic trends and synergistic application opportunities
To synthesize content across full texts, LLM models were applied to papers individually because combining full text of all papers in each cluster would both be too large and too noisy for the model.
% , this part of the analysis was performed on papers individually. 
The systematic process of cleaning each paper, identifying best-performing LLM models, extracting key information, and synthesizing findings is as follows.

After cleaning and preparing each paper, they were passed to the model using the same approach as the abstract-level analysis. 
Using the first \texttt{12,000} characters of each paper to foster smooth processing and improve model success \cite{Beltagy2020}, this work evaluated both \texttt{LLaMA-3 8B} and \texttt{Mistral-7B Instruct} for comparative purposes. 
% \textcolor{red}{Upon performing experiments with both, \texttt{Mistral-7B Instruct} outperformed \texttt{LLaMA-3 8B} when it came to extracting structured information, staying literal and grounded in the text, following strict extraction instructions, and producing structured outputs, which is generally what \texttt{Mistral} models are known for. }
Prompt engineering was a challenge when working with full papers too due to the precision requested in both extraction and output, but both the model choice and the clear and concise guidelines assisted in this process. 
To address the limited LCA knowledge in open-source pre-trained models, the prompt was designed to include both a brief introduction to the four main LCA stages and a list of common LCIA methodologies. 
The prompt then requested the model to parse each paper and output exactly seven lines containing the LCA stage, LCIA method, application area, AI/ML task, AI/ML technology, impact metrics, and claimed benefit. 
The model was instructed to output \texttt{None} for any field not addressed in the paper. 
The end result contained a clean database of 209 papers and their respective LLM outputs, all properly structured.

% Although the LLM extracted all necessary information successfully, it was critical to create categorical labels that would be easy to comprehend for visualization and further analysis. 
Although the LLM extracted all necessary information successfully, in order to enable quantitative visualization and comparative analysis, the extracted annotations were then standardized through an additional LLM-based labeling step via \texttt{Mistral-7B Instruct}.
% For that reason, the previously generated database of LLM outputs went through yet another round of adjustments via \texttt{Mistral-7B Instruct}. 
This time, however, the prompt requested the model to read through previously generated annotations and generate clean labels from a provided lists (e.g., for the AI/ML category: \texttt{ANN, SVM, LLM, Decision Trees, Reinforcement Learning, PCA, Regression}, or \texttt{Other}). 
In the end, additional basic cleaning was applied to ensure uniformity across all documents for easier comprehension and visualization.

%%%% Results %%%%%%%%%%%%%%%%%%%%%%%%%%%%%%%%%%%
\section{Results} \label{sec:results}
% "and discussion"

% By using sentence embeddings, dimensionality reduction, and hierarchical density-based clustering, as discussed above, 209 papers were successfully grouped without imposing predefined categories. These clusters were then analyzed using a combination of keyword extraction and LLM–assisted interpretation to characterize dominant research themes, methodological trends, and areas of overlap across the corpus. The complete articles were then analyzed to produce structured data that showcased specific methodologies and research areas addressed within the articles, allowing for clear analysis and visualization of trends. 

This section presents findings from the abstract-level clustering, term frequency analysis, full-text LLM extraction, and correspondence analysis.
Regarding the application of LLMs in the review methodology, this paper found that LLM-based and non-LLM based methods produced successful results while indicating preferable algorithmic choices for LCA-based systematic reviews. 
Specifically, both \texttt{Mistral-7B Instruct} and \texttt{LLaMA-3 8B} were effective, but \texttt{Mistral-7B Instruct} outperformed \texttt{LLaMA-3 8B} when it came to extracting structured information, staying literal and grounded in the text, following strict extraction instructions, and producing structured outputs. 
The outcomes of this review process are as follows.
% The findings of this systematic review of LCA-AI research are as follows.
% Papers in the corpus form well-defined groups and cover topics from different application areas and design methodologies. 
% With the overarching theme being \textit{reducing carbon emissions}, based on keyword extraction across clusters, all papers suggest creative AI-enabled solutions to current LCA challenges. 
% The growth in LCA research alone from the year of 2014 to 2025 is apparent, and the introduction of modern AI and ML technologies into the field has especially increased over the years.

\subsection{Topical Clusters}

\begin{figure}[t]
    \centering
  \includegraphics[width=0.5\linewidth]{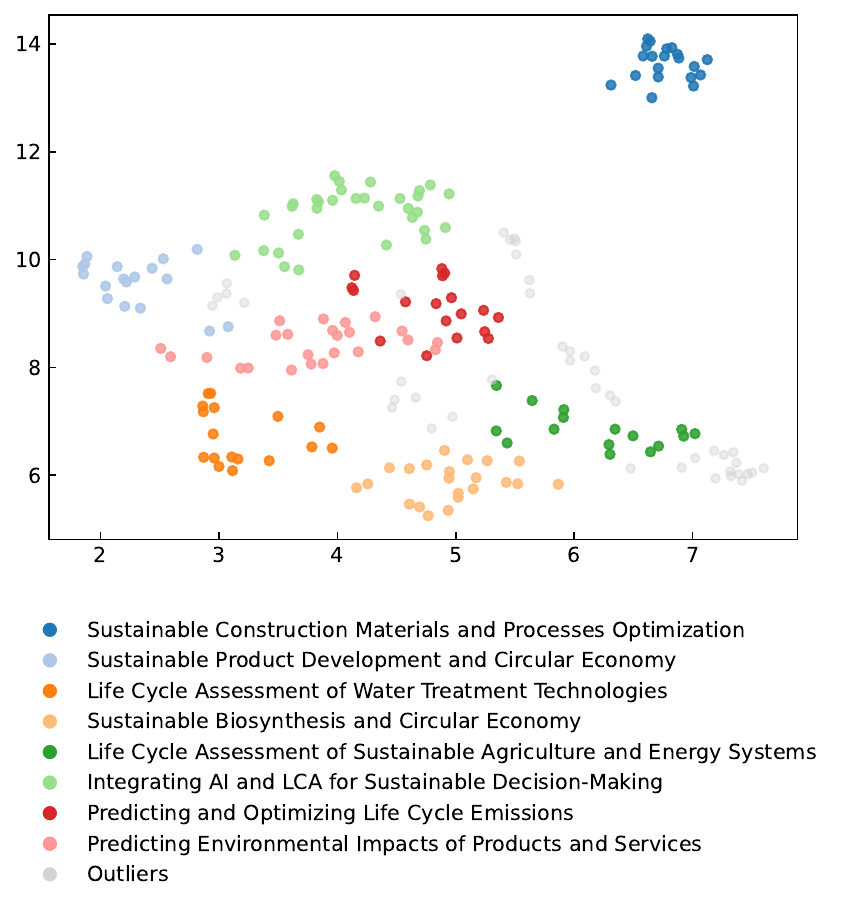}
    \caption{UMAP projections, their respective HDBSCAN-produced clusters, and LLM-generated cluster titles ($n=209$).}
    \label{fig:clusters}
\end{figure}

Embedding-based clustering revealed several coherent groupings within the AI–LCA literature. 
Figure~\ref{fig:clusters} shows the results from HDBSCAN clustering with UMAP dimensionality reduction, which produced distributions consistent with expectations.
Overall, these clusters suggest two dominant directions in the AI–LCA literature: application-driven studies (e.g., \texttt{Life Cycle Assessment of Water Treatment Technologies}) and product development and design–oriented research (e.g., \texttt{Sustainable Product Development and Circular Economy}).
% The general distribution of clusters follows expectations. 
Interestingly, an explicit AI cluster (\texttt{Integrating AI and LCA for Sustainable Agriculture and Energy Systems}) was produced even though all documents contain AI applications to an extent. 
While this result may indicate a prevalence of AI language in the cluster, the label could also be attributed to the LLM's labeling procedure itself. 
% Even though all outliers were ignored as part of the analysis, a potential alternative solution could include combining outliers with the nearest clusters.

In Fig.~\ref{fig:clusters}, it is also worth noting that although one cluster (\texttt{0: Sustainable Construction Materials and Process Optimization}) appears visually separated from the others, this separation should not be interpreted as a meaningful global distance. 
This is because UMAP prioritizes preservation of local neighborhood structure rather than absolute inter-cluster distances.
As such, cluster placement in the low-dimensional space is primarily illustrative \cite{McInnes2018}.

\begin{table}[t]
\centering
\caption{Summary of identified AI–LCA clusters including term extraction and LLM-generated approaches.
 % for validation and refinement
}
\begin{tabular}{c p{0.35\textwidth} p{0.55\textwidth}}
\hline
$C$ & Key TF-IDF Phrases & LLM-Generated Title  \\
\hline
0 & carbon emissions, recycled aggregates & Sustainable Construction Materials and Processes Optimization \\
1 & product development, sustainable manufacturing & Sustainable Product Development and Circular Economy \\
2 & wastewater treatment, human health & Life Cycle Assessment of Water Treatment Technologies \\
3 & biodiesel production, ghg emissions & Sustainable Biosynthesis and Circular Economy \\
4 & water use, resource recovery & Life Cycle Assessment of Sustainable Agriculture and Energy Systems \\
5 & energy consumption, machine learning & Integrating AI and LCA for Sustainable Decision-Making \\
6 & ghg emissions, machine learning & Predicting and Optimizing Life Cycle Emissions\\
7 & manufacturing processes, deep learning & Predicting Environmental Impacts of Products and Services \\
\hline
\end{tabular}
\label{tab:clusters}
\end{table}

After generating clusters, key terms/phrases associated with each cluster were identified to get a clearer understanding of the distribution of topics. 
Table~\ref{tab:clusters} summarizes the identified terms extracted using both an open-source LLM and standard TF-IDF approaches.  
% was queried to output a title, a brief summary, and a description of AI methodologies most prominent in the given cluster. 
For interested readers, further details on cluster topics and sample papers are provided in the Appendix. 
Evaluating these term extraction methods, it is evident that both the TF-IDF method and the LLM successfully identified key themes within each cluster. 
Although TF-IDF theme extraction was helpful, this method highlighted a lot of obviously repeating terms, such as \texttt{life}, \texttt{cycle}, \texttt{assessment}, and \texttt{environmental}.
LLMs, however, provided a more coherent and readable output without compromising details.
It is also interesting to note that the LLM found AI methodologies in all clusters, highlighting both specific methodologies used most often as well as the application areas. To understand how these clusters relate to broader research trends, we next examined overarching terms across the entire corpus.

\subsection{Overarching Terms}

\begin{figure}[t]
    \centering
  \includegraphics[width=0.5\linewidth]{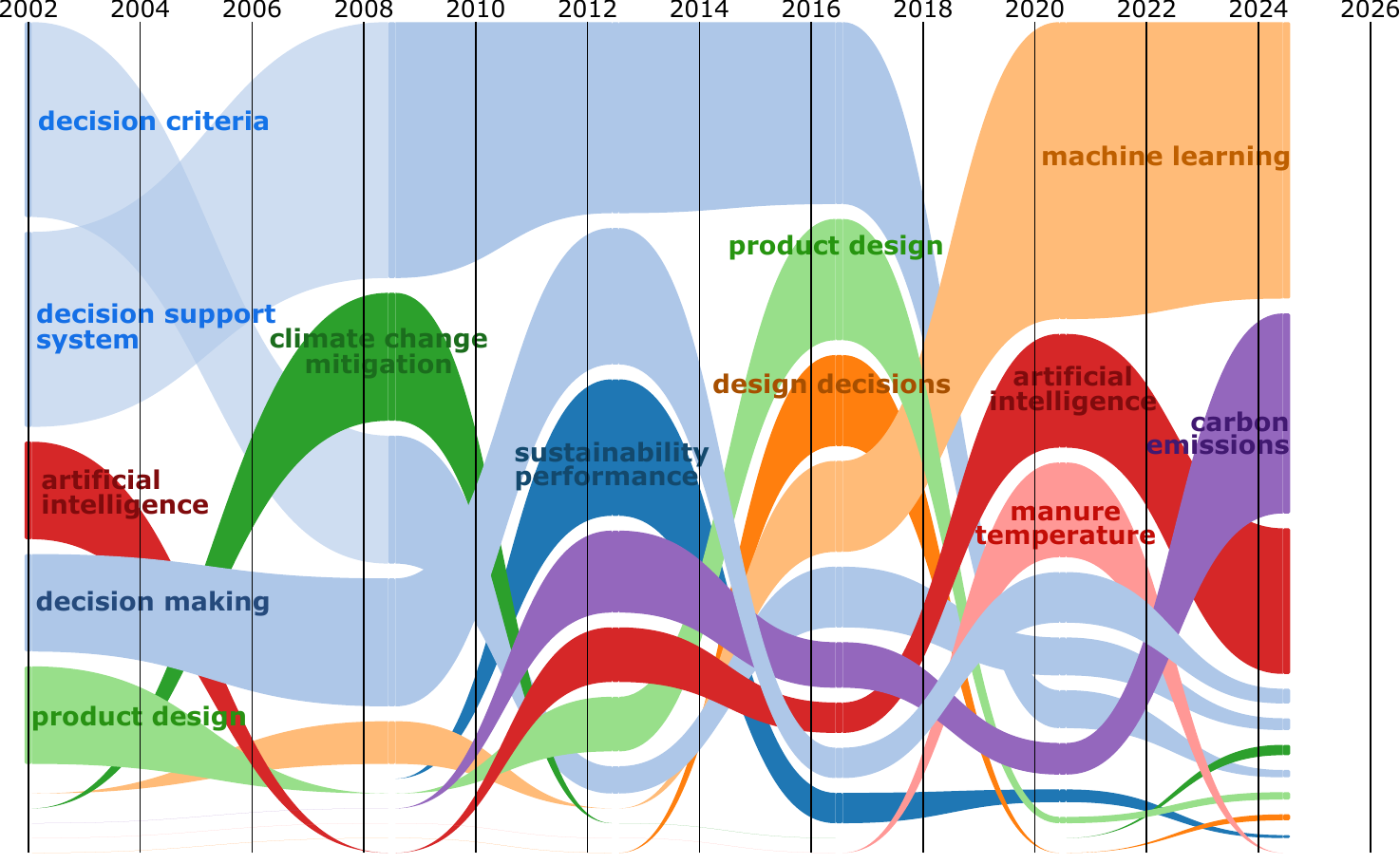}
    \caption{Evolution of the top terms in abstracts ($n=538$) over 4-year time periods with heights normalized by number of terms per period.}
    \label{fig:bumps}
\end{figure}

Supplementing the eight identified clusters, frequent terms mapped overtime indicate a general trend from decision-focused approaches to ML and other areas, as shown in Fig.~\ref{fig:bumps}. 
Most notable, ML emerges as dominant from 2018 to the present.
This indicates a strong data-centric rooting of AI-based approaches in LCA applications.
Beyond ML, carbon emissions and AI exhibit clear growth over time. 
Climate change mitigation was the second most popular term from 2006-2010, while \texttt{product design} and \emph{design decisions} peaked from 2014-2018.
These overarching terms provides a big picture view of topical trend dynamics and compliment the clusters from Fig.~\ref{fig:clusters}.

Focusing on energy and power terms specifically, several key words emerged at the intersection of LCA and AI (Tab.~\ref{tab:energy-terms}). Based on specificity, \texttt{renewable energy}, \texttt{power generation}, and \texttt{output energy} were most prevalent. 
Consistent with expectations, energy sources are important considerations for AI and LCA. 
In the future, mapping these terms with respect to AI terms, LCA terms, or clusters can inform where energy topics are concentrated or absent among the larger body of literature. 

While these trends illustrate shifts in thematic focus over time, they do not reveal which AI techniques are used. Therefore, we next analyzed full-text content to extract AI methodologies directly from papers.

\begin{table}[ht]
\caption[Table]{Most frequently occurring terms in abstracts ($n=538$) related to energy based on specificity.}\label{tab:energy-terms}
\centering{
% {%
% \tagpdfsetup{table/header-rows={1}}%

\begin{tabular}{llr}
\toprule
Term & Specificity $\chi^2$ & $f$ \\
\midrule
renewable energy & 228 & 8 \\
power generation & 215 & 8 \\
output energy & 214 & 5 \\
operational efficiency & 211 & 11 \\
biodiesel production & 209 & 3 \\
energy systems & 199 & 17 \\
energy transition & 198 & 9\\
natural gas & 196 & 9 \\
energy production & 184 & 9 \\
fuel consumption & 180 & 6 \\
energy management & 175 & 6 \\
\bottomrule
\end{tabular}
}
\end{table}

\subsection{AI Topics in LCA}

\begin{figure}[t]
    \centering
  \includegraphics[width=0.5\linewidth]{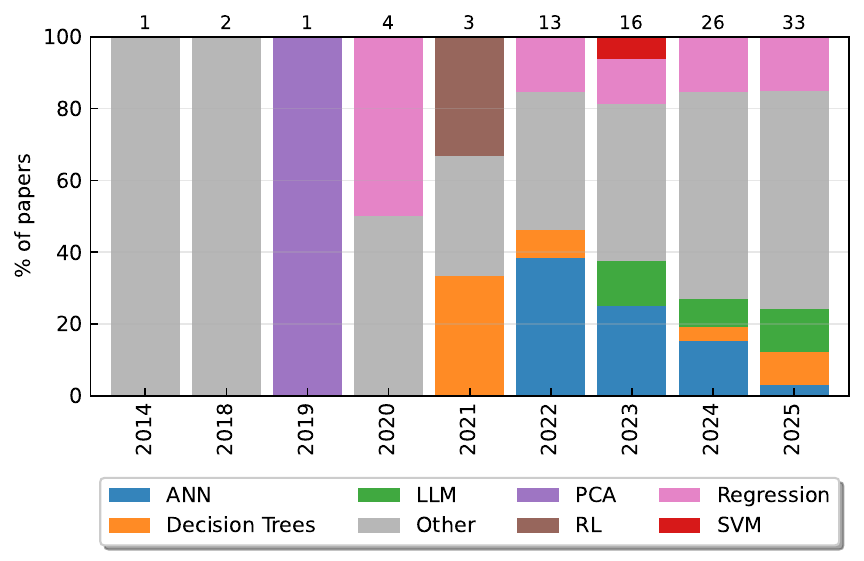}
    \caption{Distribution of AI topics over time extracted from full texts with an LLM. All papers where the LLM could not find any AI methodologies were excluded. Number of documents in each year are listed across the top.}
    \label{fig:ai_dist}
\end{figure}

In addition to analyzing clusters, full documents were parsed to extract topical information and study the distribution of research themes across years. 
Figure~\ref{fig:ai_dist} shows the distribution of core AI topics from 2014-2025. 
% As mentioned earlier, these labels were generated by using a secondary layer of LLM processing, which requested the model to convert long-form labels generated previously into concrete categorical labels. 
Although the LLM identified AI topics effectively in each paper, a large number of documents only contained generic labels, such as \texttt{Machine Learning (ML)}, \texttt{Prediction and Optimization}, \texttt{Probabilistic Analysis}, and \texttt{Data Engineering}. 
To avoid noise, these labels were grouped into \texttt{Other}. 
As shown in Fig.~\ref{fig:ai_dist}, although AI has played a role in LCA research for many years, the majority of significant AI applications have emerged in the past few years, reflecting the broader trend of increasing AI adoption across industries.
Overall, the distribution shows a shift from traditional ML methods toward more diverse and specialized AI techniques after 2020.

\subsection{LCA and AI Convergence}

\begin{figure*}[t]
    \centering
    \includegraphics[width=\linewidth]{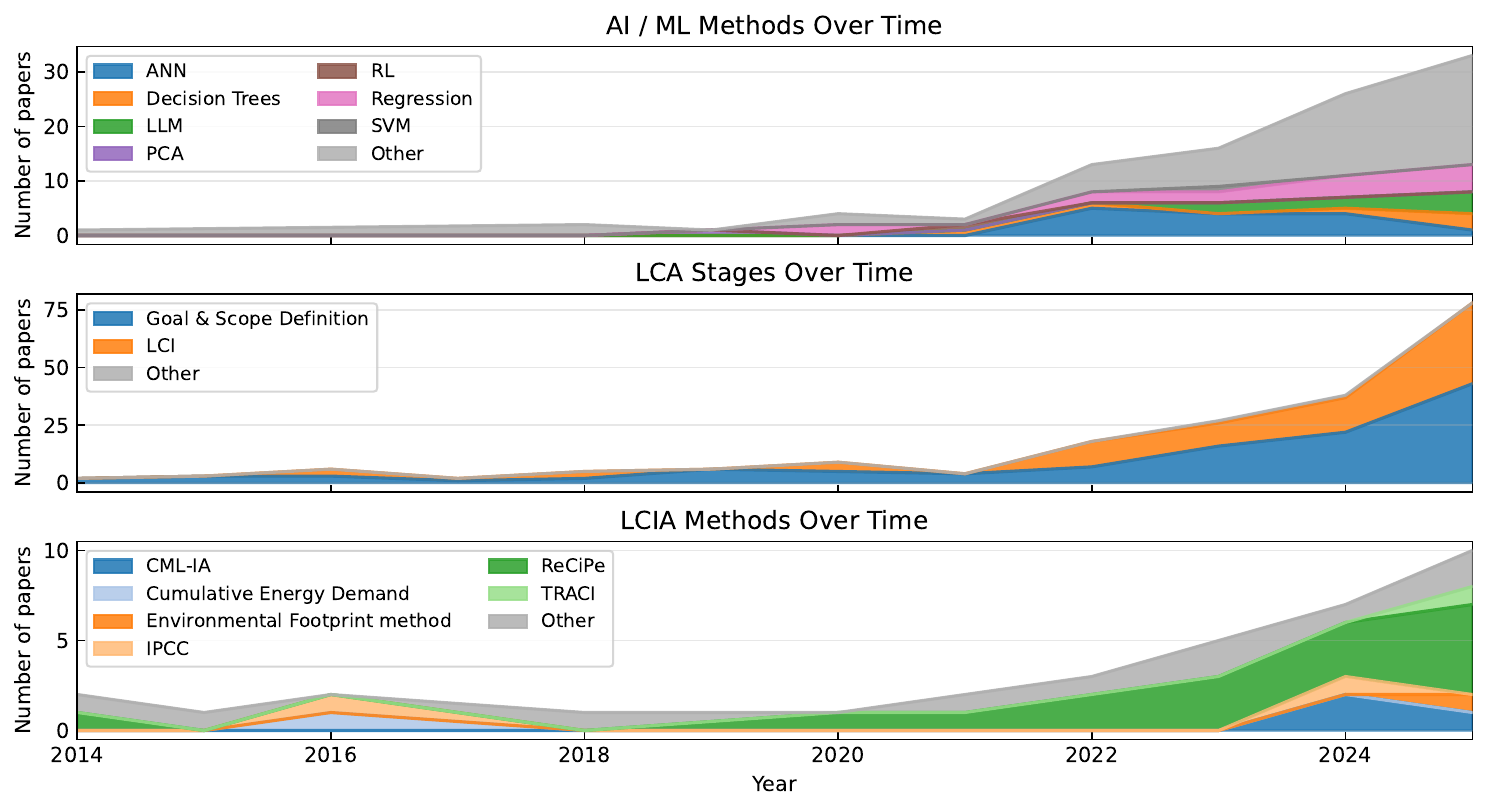}
    \caption{Distribution of AI topics, LCA stages and LCIA methodologies over time ($n=209$).}
    \label{fig:tech_counts}
\end{figure*}

Although AI topics were the focus of this study, it was equally imperative to identify trends in LCA stages and LCIA methodologies for contextual awareness. Figure~\ref{fig:tech_counts} shows the distribution of AI methods, LCA stages, and LCIA methods from 2014-2025. 
These findings show growing attention to \texttt{Goal \& Scope Definition} and \texttt{LCI} stages of LCA.
The use of AI technologies is increasing as well, with a steady growth seen in \texttt{Regression}, \texttt{LLMs}, and other AI/ML methods. 
Similarly, LCIA methods also experience a steady increase in use, with noticeable growth in \texttt{TRACI} and \texttt{ReCiPe}, both of which represent comprehensive impact assessment methodologies.

\begin{figure}[h!]
    \centering
    \includegraphics[width=0.5\linewidth]{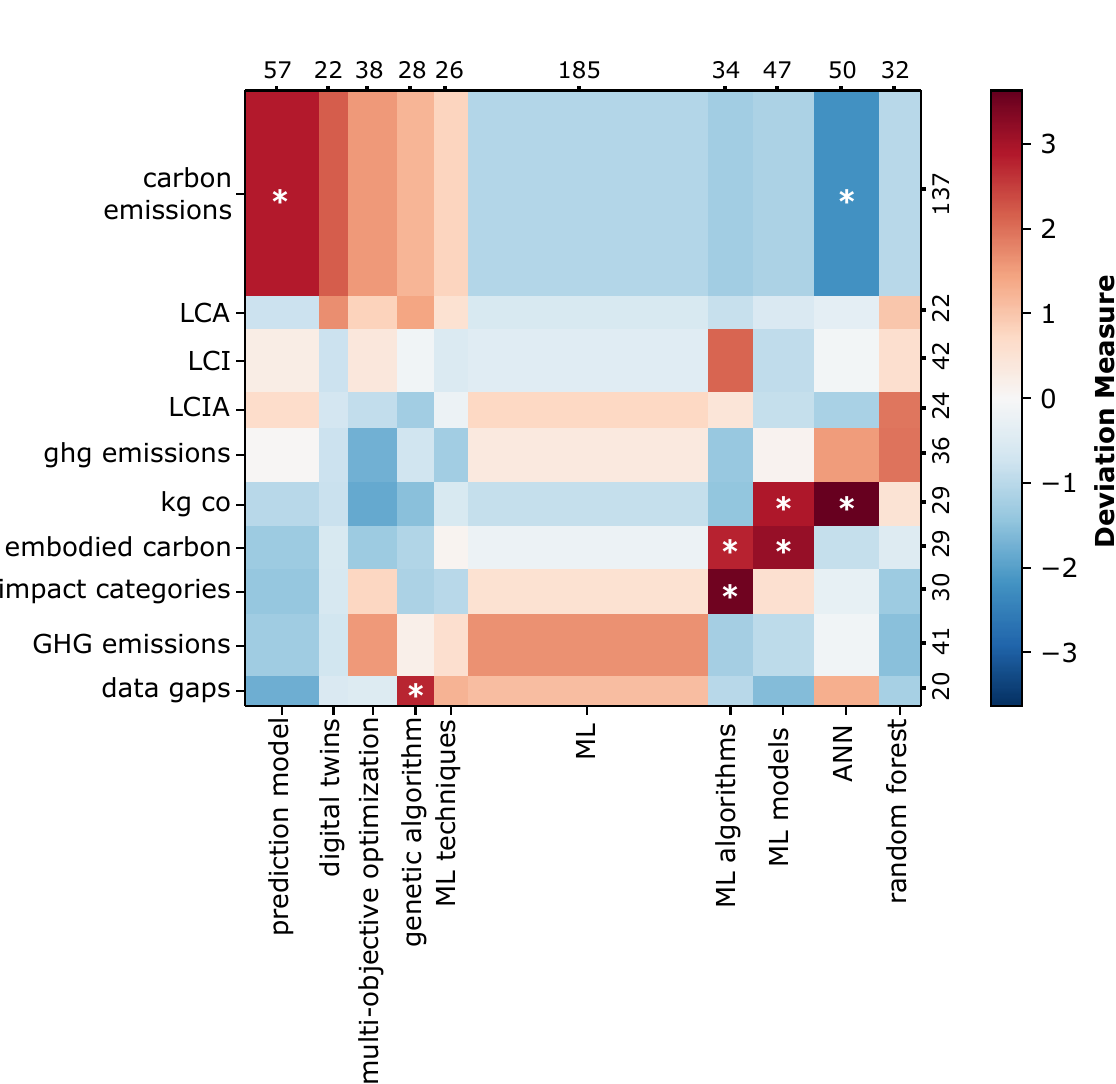}
    \caption{Contingency between AI/ML and LCA terms across abstracts ($n=538$) indicating highly statistically significant differences ($p=0.00262$) across all terms. Cells with an asterisk are significant ($p<0.05$).}
    \label{fig:contingency}
\end{figure}

Further, correspondence between AI and LCA terminology produced statistically significant differences (Fig.~\ref{fig:contingency}).
Genetic algorithms appear frequently for addressing data gaps.
Among ML-based terms, algorithms and models commonly occur alongside carbon-based terms (\texttt{kg co} and \texttt{embodied carbon}), while \texttt{prediction model} and \texttt{carbon emissions} produced strong correlations. 
Interestingly, ANNs have strong co-occurrence with \texttt{kg co} but strong disassociation with \texttt{carbon emissions}; 
this indicates the possibility of terminological differences between nuanced domains.
Overall, these findings reveal statistically highly significant correlations between AI and LCA terms ($p=0.00262$).
From these findings, it appears that the research community has started to establish trends for when and where various AI methods are adopted in LCA applications.  
Collectively, these findings reveal a rapidly evolving AI–LCA research landscape characterized by increasing methodological diversity, stronger linkages between AI methods and specific LCA stages, and a noticeable shift toward LLM-enabled analyses. 
The clustering, term mapping, and correspondence analyses all point to consistent thematic structures across the literature, confirming the robustness of the LLM-assisted review methodology.

%%%%% Discussion %%%%%%%%%%%%%%%%%%%%%%%%%%%%%%%
% \section{Discussion}
% \textcolor{red}{[wait]}

%%%%% Conclusion %%%%%%%%%%%%%%%%%%%%%%%%%%%%%%%

\section{Conclusion} \label{sec:conclusion}
This work examined how AI has been integrated into LCA research, how these applications have evolved, and how LLM-based methods can support systematic reviews in the LCA domain. 
Through a multi stage approach combining  metadata screening, embedding-based clustering, UMAP dimensionality reduction,  HDBSCAN clustering, and LLM-driven interpretation, we mapped the structure of LCA-AI literature and identified key methodological and thematic patterns. 
Full-text documents were further analyzed using two rounds of LLM-assisted data synthesis, creating a clean breakdown of technological advancements and enabling a clear and reproducible synthesis of technological developments across the field. 
Taken together, the results demonstrate a rapidly expanding use of AI within LCA, spanning many diverse sectors and methodological stages. 
The significant correlations between AI-related and LCA-related terminology highlight a stronghold of data-centric ML methods, while LLMs and other emerging AI algorithms fill important niches. 
Looking ahead, this work opens the door to several promising directions; future efforts will extend full-text analysis to all screened papers ($n=538$), incorporate full-text embeddings into the clustering framework, perform a deeper dive into AI/ML methodologies, and evaluate additional LLM-assisted architectures and prompting strategies to supplement existing results. 
More broadly, the findings illustrate how LLM-assisted analysis can enhance scalability, reproducibility, and depth in literature reviews for rapidly evolving research domains, offering a foundation for more efficient, data driven synthesis across the LCA community.

%%%%% Acknowledgments %%%%%%%%%%%%%%%%%%%%%%%%%%

% \section*{Data Availability} \label{sec:metadata}
% A static URL to the literature review metadata included in this analysis will be added after peer review.

\section*{AI Statement}

This paper was written in-full by the authors, including initial drafts and editing. 
\texttt{LLaMA-3 8B} and \texttt{Mistral-7B Instruct} were used to extract natural language from existing original research in this review. 
AI was not used to create original research. 
The authors have verified and take full responsibility for all outputs from LLM and AI tools used herein.  

\section*{Acknowledgments}
This research was supported by the U.S. National Science Foundation under Grant CBET-2501735. 
Any opinions, findings, and conclusions or recommendations expressed in this material are those of the authors and do not necessarily reflect the views of the National Science Foundation.

\bibliographystyle{IEEEtran} 
\bibliography{references}

%%%  APPENDICES  %%%%%%%%%%%%%%%%%%%%%%%%%%%%%%%%
\appendix

\section{Cluster Details}\label{sec:appendix}

Table~\ref{tab:cluster-details} details the topics and descriptions of all eight clusters.
The top two ranked citations are provided as representative papers for each cluster.
Please see Tab.~\ref{tab:clusters} for frequent terms corresponding to each cluster $C$.

% \FloatBarrier
\begin{table*}[h!] %[!t]
\centering
\captionof{table}{Summary of identified AI–LCA clusters including term extraction and LLM-generated approaches for validation and refinement.}
\begin{tabular}{c p{0.25\textwidth} p{0.54\textwidth} p{0.1\textwidth}}
\hline
$C$ & LLM-Generated Title & LLM-Generated AI Methods & Top Ranked Citations\\
\hline
0 & Sustainable Construction Materials and Processes Optimization & Machine learning models, such as neural networks and genetic algorithms, are used to predict and optimize various performance metrics, including energy consumption, carbon emissions, and cost & \cite{de_Paula_Salgado_2025, Katish_2024}\\
1 & Sustainable Product Development and Circular Economy & Machine learning algorithms are used in some papers to analyze data and predict outcomes, while others employ AI-powered decision support systems for optimizing resource efficiency & \cite{Exner_2025, Larrea_Gallegos_2025}\\
2 & Life Cycle Assessment of Water Treatment Technologies & Machine learning models are used to predict chemical toxicity and expand species sensitivity distributions, while deep neural networks enhance environmental impact assessment accuracy for constructed wetlands & \cite{Gjedde_2024, Song_2021} \\
3 & Sustainable Biosynthesis and Circular Economy & Machine learning (ML) techniques, such as transformer-based language models and natural gradient boosting, are used to optimize processes, predict yields, and analyze LCA data & \cite{Ascher_2022, Hajabdollahi_Ouderji_2023} \\
4 & Life Cycle Assessment of Sustainable Agriculture and Energy Systems & Machine learning models such as random forest regression, Gaussian kernel regression, and neural networks are used to predict missing data, optimize datasets, and improve the accuracy of life cycle climate emissions predictions & \cite{Ratledge_2022, Foschi_2025} \\
5 & Integrating AI and LCA for Sustainable Decision-Making & Papers in this cluster employ AI/ML techniques such as generative design, predictive modeling, and semantic interoperability to analyze and improve the environmental performance of buildings, infrastructure, and manufacturing systems & \cite{El_Hafdaoui_2023, Mart_nez_Rocamora_2021} \\
6 & Predicting and Optimizing Life Cycle Emissions & Machine learning algorithms, such as random forest and gradient boosting, are used to predict LCA outcomes, optimize routes, and identify key factors influencing energy consumption and emissions & \cite{Ranpara_2025, Symonds_2025} \\
7 & Predicting Environmental Impacts of Products and Services & Machine learning algorithms, such as random forest, neural networks, and extreme gradient boosting, are used to predict LCA data and improve prediction accuracy & \cite{Nwagwu_2025, Boedijanto_2024} \\
\hline
\end{tabular}
\label{tab:cluster-details}
\end{table*}

\end{document}